\documentclass[runningheads]{llncs}
\usepackage[T1]{fontenc}
\usepackage{graphicx,verbatim}
\usepackage{tabularx}
\usepackage{multirow}
\usepackage{booktabs}  
\usepackage{amsmath}
\usepackage{hyperref}
\begin{document}
\title{Are ECGs enough? Deep learning classification of pulmonary embolism using electrocardiograms}
\titlerunning{Are ECGs enough? Deep learning classification of
PE}
%
\begin{comment}  %% Removed for anonymized MICCAI 2025 submission
\author{First Author\inst{1}\orcidID{0000-1111-2222-3333} \and
Second Author\inst{2,3}\orcidID{1111-2222-3333-4444} \and
Third Author\inst{3}\orcidID{2222--3333-4444-5555}}
%
\authorrunning{F. Author et al.}
\institute{Princeton University, Princeton NJ 08544, USA \and
Springer Heidelberg, Tiergartenstr. 17, 69121 Heidelberg, Germany
\email{lncs@springer.com}\\
\url{http://www.springer.com/gp/computer-science/lncs} \and
ABC Institute, Rupert-Karls-University Heidelberg, Heidelberg, Germany\\
\email{\{abc,lncs\}@uni-heidelberg.de}}

\end{comment}

\author{João D.S. Marques
\and
Arlindo L. Oliveira
}  %% Added for anonymized MICCAI 2025 submission
\authorrunning{João D.S. Marques et al.}
\institute{INESC-ID / Instituto Superior Técnico \\
    \email{\{joao.p.d.s.marques, arlindo.oliveira\}@tecnico.ulisboa.pt}}

\maketitle              % typeset the header of the contribution
\begin{abstract}
Pulmonary embolism is a leading cause of out of hospital cardiac arrest that requires fast diagnosis. While computed tomography pulmonary angiography is the standard diagnostic tool, it is not always accessible. Electrocardiography is an essential tool for diagnosing multiple cardiac anomalies, as it is affordable, fast and available in many settings. However, the availability of public ECG datasets, specially for PE, is limited and, in practice, these datasets tend to be small, making it essential to optimize learning strategies. In this study, we investigate the performance of multiple neural networks in order to assess the impact of various approaches. Moreover, we check whether these practices enhance model generalization when transfer learning is used to translate information learned in larger ECG datasets, such as PTB-XL, CPSC18 and MedalCare-XL, to a smaller, more challenging dataset for PE. By leveraging transfer learning, we analyze the extent to which we can improve learning efficiency and predictive performance on limited data.

% Authors must provide keywords and are not allowed to remove this Keyword section.

\end{abstract}
%
%
%

%Start of the paper
\section{Introduction}

Cardiovascular diseases (CVDs) represent a significant global health challenge, being one of the leading causes of mortality and a significant contributor to the reduction in quality of life \cite{perigo_doencas_cardiacas}. Among the most severe conditions, pulmonary embolism (PE) is the third most common CVD after stroke and heart attack \cite{pe_danger_incidence}, accounting for 2–9\% of all causes of out-of-hospital cardiac arrest \cite{cardiac_arrest_pe}. Due to its typically poor prognosis, fast diagnosis and treatment are critical. Although computed tomography pulmonary angiography (CTPA) is often required to confirm the diagnosis of PE, it is not feasible for all patients and may even be inaccessible in certain healthcare facilities \cite{diagnostico_pe}.

Electrocardiography (ECG) is a widely used diagnostic tool in both pre-hospital and hospital settings due to its affordability, speed, and accessibility \cite{Livro_ecgs}. While improvements in ECG-based scoring systems \cite{Daniel_score,PegD_score,Wells_score} have enhanced diagnostic capabilities, they remain time-consuming and are not routinely applied in clinical practice. Moreover, the non-specific nature of PE symptoms, which frequently overlap with other CVDs, makes ECG-based detection particularly challenging \cite{diagnostico_pe}.

The rise of foundational AI models is having an impact on various domains, including medicine \cite{gpt_medical_capabilities,med_gemini}, yet anomaly detection using ECGs remains highly specialized, posing challenges for model adaptation. Although deep learning offers promising potential in this area, its effectiveness is limited by the lack of publicly available ECG datasets, particularly for PE. The few existing private datasets are typically small, containing less than 1,000 samples, and highly imbalanced, which poses significant challenges for model training and generalization.

Given these difficulties, this study aims to establish a baseline performance for deep learning models applied to PE detection using ECGs and to evaluate the impact of transfer learning in settings characterized by limited and imbalanced datasets. Our results show that, despite data scarcity, transfer learning can significantly enhance detection performance and support more effective clinical decision-making.

All code, architectures and hyperparameters are publicly available at \url{https:github.com/joaodsmarques/Are-ECGs-enough-Deep-Learning-Classifiers}.

\section{Background}

The application of deep learning to ECG analysis is not new. Multiple studies have explored this field over the years \cite{miccai_ecg_paper1,Transformer_ECG}. However, progress remains challenging, largely due to the limited availability of publicly accessible datasets. 

A 1D-CNN is often the standard approach for ECG classification, as convolutional layers effectively capture local features from the signal, achieving high performance in several classification tasks \cite{ECG_class_justCNN}. However, these architectures struggle to recognize long-range dependencies. Recurrent neural networks (RNNs), particularly LSTMs and GRUs \cite{Comparison_gru_Lstm}, are well-suited for capturing the temporal dependencies of ECGs. As demonstrated by Zahra Ebrahimi et al. \cite{from_cnn_to_RNN}, both 1D-CNNs and RNN-based models can extract relevant features for ECG classification.

Recently, Narotamo et al. \cite{paper_margarida_silveira} explored multiple deep learning architectures for detecting cardiac anomalies using the PTB-XL dataset. In their 1D experiments, they found that CRNNs, using GRU and LSTM, offer an effective balance between feature extraction and capturing temporal dependencies.

Beyond traditional architectures, new transformer-inspired models have been proposed for ECG analysis. For example, ECG-Mamba \cite{ecg_mamba} achieved an F1-score of 75\% on PTB-XL and 79\% on CPSC18, demonstrating the potential of new architectures in ECG classification.

In the context of PE classification, Somani et al. demonstrated that ECGs provide valuable diagnostic information for PE detection and can enhance the performance beyond what is achievable using CTPA alone \cite{somani_pe_ecgs}.

In this work, we investigate several deep learning architectures applied to different ECG datasets to evaluate their effectiveness in learning directly from raw ECG signals. We then assess their performance in the context of PE detection, aiming to determine whether ECGs can contribute to improved detection in scenarios where other diagnostic tools are unavailable.

\section{Methodology}

Publicly available datasets for ECG anomaly detection are limited, and those that exist are often small and imbalanced. To maximize their utility, our experiments are guided by the following key questions:
\begin{itemize}
    \item \textbf{What performance can be expected when training from scratch?} We train multiple models on two datasets to identify best practices under these settings.
    \item \textbf{How effective is transfer learning?} We investigate whether transfer learning from three larger datasets improves performance in a smaller PE detection setting.
    \item \textbf{Can our models outperform clinical approaches for PE detection?} We evaluate whether our best-performing model can surpass existing clinical procedures.
\end{itemize}

\subsection{Datasets}
We begin this study by focusing on 3 ECG datasets: PTB-XL \cite{wagner_ptb-xl_2020}, the China Physiological Signal Challenge 2018 (CPSC18) \cite{cpsc18_2018}, and MedalCare-XL \cite{medalcare_dataset}. Insights gained from these datasets are then applied to the PE using the PE-HSM dataset \cite{VALENTESILVA2023643}. 

The selection of datasets is based on the objective of training on the largest and most complete publicly available dataset, PTB-XL, while using CPSC18 as a comparative benchmark due to its relatively large size. In addition, we assess the utility of synthetic ECGs through the MedalCare-XL dataset. Finally, we evaluate whether the insights gained from these datasets generalize to the PE-HSM setting. All selected datasets share key characteristics: 500 Hz sampling frequency, 12-lead ECG recordings, and adaptable signal lengths, ensuring consistency across experiments. An overview of these datasets is provided in Table~\ref{tab:datasets}.

\begin{table}[h]
    \centering
    \caption{Overview of the datasets used for each selected task.}
    \renewcommand{\arraystretch}{1.1} % Adjust row height for better spacing
    \begin{tabular}{p{1.7cm} p{2cm} p{1.8cm} p{2.5cm} p{1.9cm} p{1.7cm}}
        \toprule
        \textbf{Dataset} & \textbf{Availability} & \textbf{Task} & \textbf{Classification} & \textbf{Samples} & \textbf{Length} \\
        \midrule
        PTB-XL  & Public  & Diagnostic & Multi-label (5)  & 21,837 & 10s \\
        CPSC18  & Public  & Arrhythmia & Multi-label (9)  & 6,877  & 6–60s \\
        MedalCare & Public  & Form & Multi-class (8)  & 16,900  & 10s \\
        PE-HSM  & Private & Acute PE   & Binary (1)       & 927    & 10s \\
        \bottomrule
    \end{tabular}
    \label{tab:datasets}
\end{table}

\begin{figure}[htbp]
    \centering
    \includegraphics[width=0.95\linewidth]{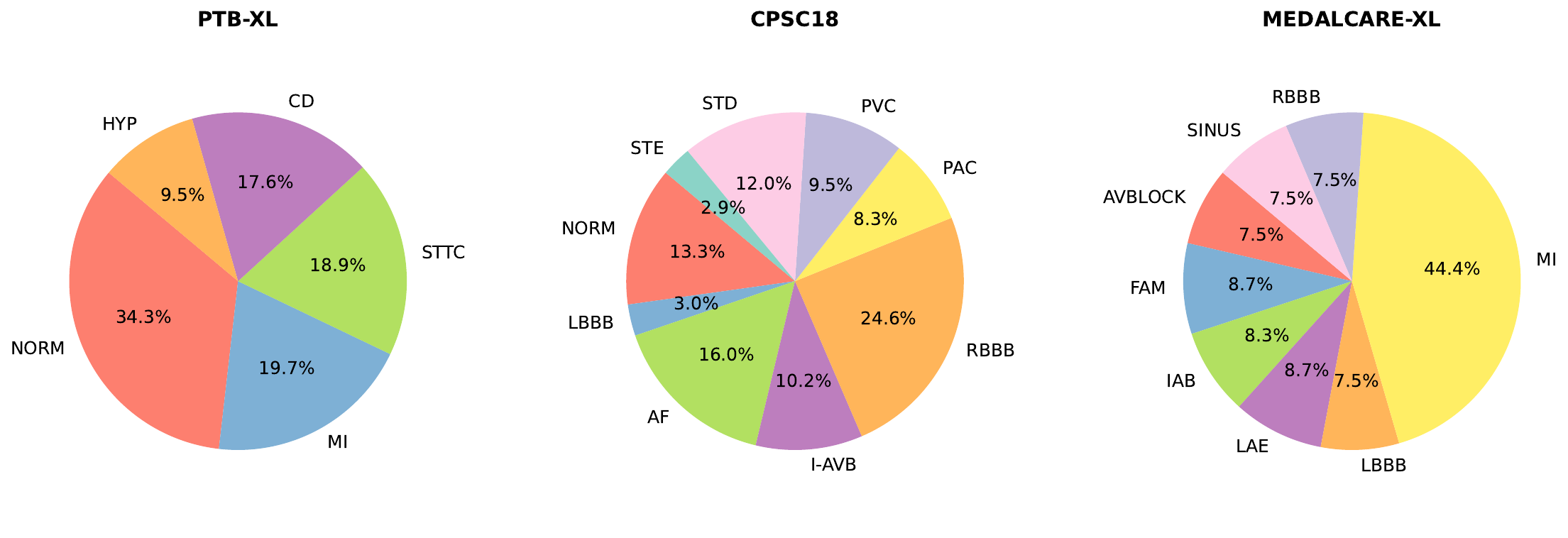}
    \caption{Class label distributions for the PTB-XL (\textbf{left}), CPSC18 (\textbf{middle}), and MedalCare-XL (\textbf{right}) datasets.}
    \label{fig:class_distribution_public}
\end{figure}

In the PTB-XL dataset, we focus on the five diagnostic superclasses instead of the 44 subclasses, as fine-tuning on all subclasses is constrained by severe class imbalance. This imbalance can bias the learned representations toward dominant classes, ultimately affecting transfer learning performance. Figure~\ref{fig:class_distribution_public} shows the class distributions across all experimental settings.

%\begin{figure}[htbp]
%    \centering
    % First image
%    \begin{minipage}{0.4\textwidth} % Adjust width
%        \centering
%        \includegraphics[width=\linewidth, trim=500 250 500 250, clip]{images/PTBXL Super class.png}
        %\caption{PTB-XL superclasses.}
%        \label{fig:class_ptbxl}
%    \end{minipage}
%    \hspace{0.2cm}
    % Second image
%    \begin{minipage}{0.4\textwidth} % Adjust width
%        \centering
%        \includegraphics[width=\linewidth, trim=500 250 500 250, clip]{images/cpsc18.png}
        %\caption{CPSC18 classes.}
%        \label{fig:cpsc_class}
%    \end{minipage}
    
%    \caption{PTB-XL class distribution (\textbf{Left}) and CPSC18 class distribution (\textbf{Right}).}
%    \label{fig:class_distribution_public}
%\end{figure}

{\bfseries Dataset splits.} For PTB-XL, we followed the protocol proposed by the original authors \cite{wagner_ptb-xl_2020}, using fold 9 for validation and fold 10 for testing. For CPSC18, we created 10 folds, ensuring representation of all classes in each one, and performed cross-validation. For MedalCare-XL, we merged all class-specific folders belonging to the same split into unified sets, and used these for model development. For PE-HSM, we adopted the original training and test partitions provided by the authors, comprising 824 training samples (222 positive and 602 negative) and 103 test samples (39 positive and 64 negative).

\subsection{Pre-processing and data augmentation}

\begin{figure}[htbp]
    \centering
    \includegraphics[width=0.95\linewidth, trim=375 275 375 275, clip]{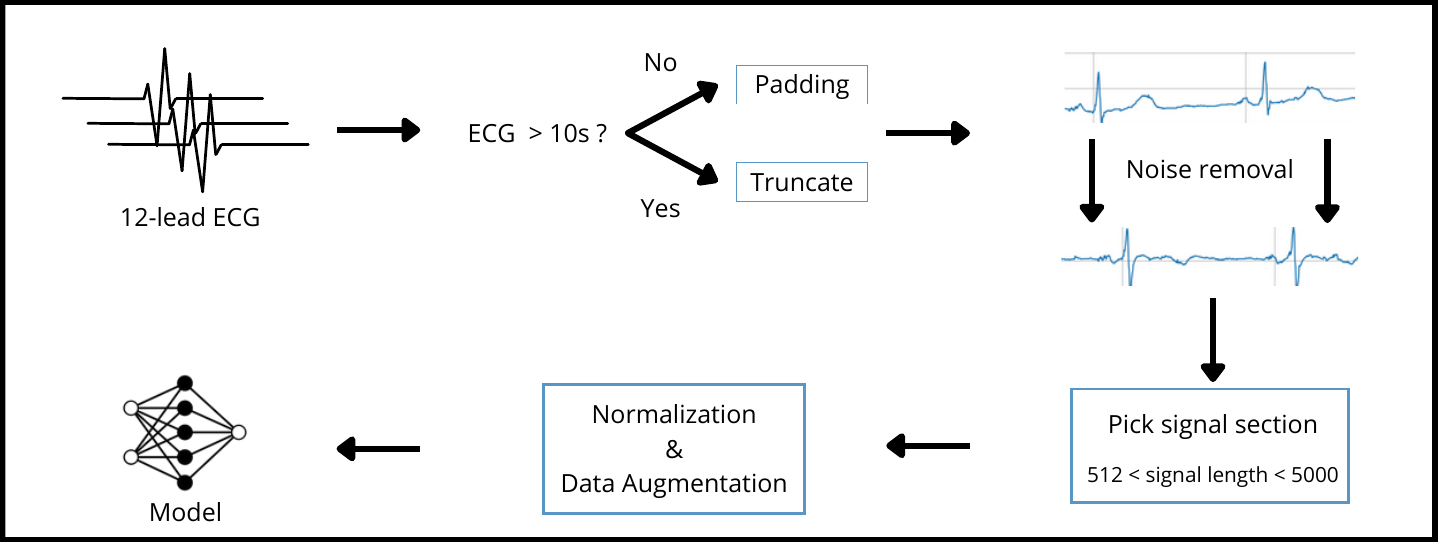}
    \caption{Pre-processing pipeline.}
    \label{fig:pipeline}
\end{figure}

Pre-processing is crucial, as it helps the models focus on relevant signal features and ignore noise \cite{ecg_pre-processing_2024}. While specialized algorithms exist for detecting and enhancing specific waveforms, such as R-peaks \cite{sadhukhan_r-peak_2012}, we adopt a simpler approach to assess the performance of neural networks from raw ECG signals. Our pre-processing pipeline consists of four key steps: noise removal (addressing power line interference and baseline drift), sequence length selection, normalization, and data augmentation. The full pipeline is illustrated in Figure \ref{fig:pipeline}.

To mitigate noise, we first applied a low-pass filter at 50 Hz to remove power line interference, followed by wavelet decomposition \cite{wavelet_decomposition}. However, applying this method during training, as required for randomized data augmentation, proved to be computationally expensive. To address this, we instead employed a second-order Butterworth bandpass filter with a high-pass cutoff at 1 Hz for baseline wander suppression and a low-pass cutoff at 45 Hz, which produced similar output quality while being more computationally efficient \cite{paper_margarida_silveira}.

For sequence length selection, in order to extract a segment from each ECG lead, we first randomly select a starting point $s$, ensuring that the extracted segment remains within the maximum available length. This process is defined in Equation \ref{eq:start_point}.

\begin{equation}
s \sim \mathcal{U}(0, m - l), \quad \text{with} \quad s + l \leq m 
\label{eq:start_point}
\end{equation}

Note that $s$ is the randomly chosen starting index, $l$ is the desired segment length, and $m$ is the maximum length of the ECG signal (5000 samples). This ensures that the segment length is consistent across leads and does not exceed the available data range. Sequences exceeding 5000 samples (maximum available length) were truncated, while shorter sequences were zero-padded. A length of 2048 samples was chosen, as it provided similar performance to the longest sequence while being significantly less computationally demanding.

For normalization, we evaluated the impact of various methods, including MinMax, Z-score, Rscale, Logscale, and L2 normalization. Finally, we applied multiple data augmentation techniques, such as flip, random drop, lead drop, square pulse sum and sine sum \cite{raghu_data_augmentation_2022}.

\subsection{Model architectures}

We explored a range of architectures, from simpler models like AlexNet to more complex encoder-based networks. We started by using AlexNet \cite{alexnet} as a baseline due to its simplicity and fast training speed. Next, we investigated the VGG family \cite{vgg_family}, identifying VGG11 (with batch normalization) as the most robust contender. We then conducted experiments with ResNet architectures, where ResNet18 emerged as a strong candidate, offering a good balance between simplicity and performance, while also proving to be an effective feature extractor for other deep learning modules.

Beyond CNN-based models, we integrated GRUs and LSTMs \cite{Comparison_gru_Lstm} to capture temporal dependencies in ECG signals. To further enhance both deep feature extraction and temporal modeling, we employed CRNNs, where ResNet18 with GRU and ResNet18 with LSTM stood out as high-performing combinations. 

We also explored representing the ECG as a 2D structured signal rather than as 12 independent 1D sequences. To this end, we applied EEGNet \cite{eegnet}, a compact convolutional neural network that has demonstrated strong performance in the electroencephalography (EEG) domain.

Given the importance of attention mechanisms in capturing long-range dependencies \cite{attention_is_all}, we studied adding attention to the ResNet18, naming this architecture AttResNet.

Finally, we explored transformer-based models \cite{attention_is_all}, including transformer encoders and Residual Transformers, to assess how larger, more recent architectures perform in this setting.

\subsection{Metrics}
Given that our task is predominantly multi-label and that class distributions are imbalanced, relying only on accuracy or area under the curve (AUC) can be misleading, as models may achieve high scores by prioritizing the most abundant classes. To provide a more comprehensive evaluation, we additionally report F1-score and mean average precision (MAP). Furthermore, in clinical applications, it is critical to minimize false negatives (ensuring high sensitivity) while also keeping high specificity to confirm that detected negatives are truly negative cases. To measure this balance, we also compute the G-mean (GM).

\section{Results and Discussion}

All models were initialized using standard procedures \cite{initialization} and trained on an NVIDIA V100S GPU (32GB) installed in a DELL PowerEdge C41402 server. To determine the optimal hyperparameter configuration for each architecture, we employed Bayesian Optimization with Optuna \cite{optuna}, running 100 trials with 50 epochs each, with the objective of maximizing the F1-score. For the loss function, we analyzed both Binary Cross Entropy (BCE) with class weighting, adjusted according to dataset distribution, and Focal Loss  \cite{focal_loss} with $\gamma = 2$ and $\alpha = 0.7$. The Focal Loss proved to be the most effective, being considered essential for achieving strong results.
    
\begin{table}[h]
    \caption{Performance (\%) of multiple models trained from scratch on PTB-XL and CPSC18 datasets.}
    \centering
    \renewcommand{\arraystretch}{1.1} % Adjust row height for better spacing
    \begin{tabularx}{\textwidth}{lXXXXXXXc}
        \toprule
         & \multicolumn{4}{c}{\textbf{PTB-XL}} &  \multicolumn{4}{c} {\textbf{CPSC18}} \\
        \textbf{Models} & \textbf{Acc} & \textbf{F1} & \textbf{MAP} & \textbf{GM} & \textbf{Acc} & \textbf{F1} & \textbf{MAP} & \textbf{GM} \\
        \midrule
        AlexNet & 85.9 & 74.3 & 76.1 & 83.8 & 93.2 & 74.8 & 72.4 & 86.2  \\
        VGG11\_bn & 86.2 & 74.1 & 77.3 & 83.2 & 94.0 & 75.0 & 70.9 & 86.1  \\
        ResNet18 & 86.8 & 75.1 & \textbf{78.9} & 83.9 & 93.9 & 75.3 & 72.0 & 87.2  \\
        EEGNet & 83.4 & 62.2 & 70.1 & 75.4 & 91.5 & 54.1 & 59.4 & 74.0  \\
        CRNN\textsubscript{LSTM} & 87.1 & 75.8 & 76.7 & 84.6 & \textbf{95.1} & 79.3 & 76.8 & 88.4  \\
        CRNN\textsubscript{GRU} & \textbf{87.2} & \textbf{76.1} & 76.7 & 84.9 & \textbf{95.1} & \textbf{79.5} & \textbf{78.3} & 88.6  \\
        AttResNet & 87.0 & 75.6 & 77.1 & \textbf{85.0} & 94.8 & 78.5 & 76.8 & 88.1  \\
        Transformer & 75.0 & 50.7 & 44.0  & 64.9 & 87.9 & 57.5 & 51.5 & 74.4  \\
        ResTransformer & \textbf{87.2} & 75.7 & 76.6 & 84.0 & 94.8 & 78.8 & 77.5 & \textbf{89.0}  \\
        \bottomrule
    \end{tabularx}
    \label{tab:models_on_public}
\end{table}

Given their clinical relevance and frequent use in the literature, PTB-XL and CPSC18 served as benchmarks for comparing model architectures. The best results obtained from training the models from scratch are summarized in Table~\ref{tab:models_on_public}. The optimized architectures include the following configurations: (a) CNNs: logscale normalization and learning rate (lr) $\approx$ 0.0002; (b) CRNNs: ResNet18, L2 normalization, 2 recurrent layers, 256 hidden size and lr $\approx$ 0.0005; (c) AttResNet: ResNet18 combined with multihead attention layer, 512 embedding dimension, 4 heads, L2 normalization and lr $\approx$ 0.0002; (d) Transformer: 1 convolutional layer to increase dimensions, 4 transformer encoder layers, 512 embedding dimension, 4 heads, zscore normalization and lr $\approx$ 0.0001; (e) ResTransformer: ResNet18, transformer encoder, 512 embedding dimension, 4 heads, zscore normalization and lr $\approx$ 0.0002; (f) EEGNet: zscore normalization and lr $\approx$ 0.0015.

Based on the results obtained in Table \ref{tab:models_on_public}, we chose 4 models to apply to the PE-HSM dataset: the best CNN (ResNet18), the best CRNN (CRNN\textsubscript{GRU}), the Attention ResNet (AttResNet) and the ResTransformer. The performance of the four models was evaluated under both transfer learning and training from scratch settings. In the transfer learning scenario, performance improved when all weights were fine-tuned, rather than updating only the final layer. Results for these settings are shown in Table~\ref{tab:models_on_hsm}.

\begin{table}[h]
    
    \caption{Performance (\%) of the best models on the PE-HSM dataset.}
     \label{tab:models_on_hsm}
     \centering
    \renewcommand{\arraystretch}{1} % Adjust row height for better spacing
    \begin{tabularx}{\textwidth}{lXXXXc}
        \toprule
        \multicolumn{6}{c}{\textbf{PE-HSM}}\\
        \textbf{Models} & \textbf{Pretrain}  & \textbf{Acc} & \textbf{F1} & \textbf{MAP} & \textbf{GM}\\
        \midrule
        \multirow{3}{*}{ResNet18} & None & 58.3 & 56.6 & 42.2 & 59.6 \\
         & PTB-XL  &  \textbf{63.1} & 61.2 & 50.8 & \textbf{64.9} \\
         & CPSC18 &  63.0 & \textbf{62.4} & 55.1 & 62.7  \\
         & MedalCare &  59.2 & 58.3 & \textbf{57.0} & 59.3  \\
        \midrule
        \multirow{3}{*}{CRNN\textsubscript{GRU}} & None & 66.0 & 61.9 & 48.2 & 70.1\\
        &  PTB-XL & \underline{\textbf{68.9}} & \textbf{65.4} & 49.6 & \textbf{72.7} \\
        & CPSC18 &  67.0 & 63.2 & 54.1 & 70.81\\
        & MedalCare &  66.0 & 61.9 & \underline{\textbf{57.2}} & 70.1  \\
        \midrule
        \multirow{3}{*}{AttResNet} & None &  64.1 & 60.2 & 52.4 & 67.8\\
        & PTB-XL  &  \textbf{68.0} & \textbf{64.1} & 53.8 & 71.9 \\
         & CPSC18 &  64.1 & 63.4 & 55.7 & 63.5  \\
         & MedalCare & \textbf{68.0} & 60.0 & \textbf{56.8} & \underline{\textbf{74.2}}  \\
         \midrule
        \multirow{3}{*}{ResTransformer} & None & 66.0 & 57.6 & 49.3 & \textbf{72.3} \\
        & PTB-XL  & 63.1 & 60.8 & 50.3 & 61.8  \\
        & CPSC18 & \textbf{68.0} & \underline{\textbf{66.7}} & 51.9 & 68.7 \\
        & MedalCare & 64.1 & 59.1 & \textbf{55.6} & 68.6  \\
        \bottomrule
    \end{tabularx}

\end{table}

While no model dominates across all metrics, the results offer valuable insights. Transfer learning increases performance on PE detection, as evidenced by improvements in all pretrained settings. CRNNs offer the best balance between ease of training and performance, while ResNet18 proves to be a strong and stable feature extractor. Attention modules benefit greatly from feature extractors, whereas transformers struggle due to limited training data.

In the PTB-XL pretraining setting, AttResNet seems close to the CRNN\textsubscript{GRU}, suggesting that with more data, attention mechanisms may outperform RNNs. Interestingly, ResTransformer achieves better results when pretrained on CPSC18 than on PTB-XL, likely due to CPSC18’s focus on arrhythmias - including a class for Right Bundle Branch Block (RBBB), a key ECG indicator of pulmonary embolism, which demonstrates effective domain adaptation.

Pretraining on MedalCare-XL yields the highest MAP across all models, improving model precision across all classes. Moreover, models generally benefit from pretraining on this dataset, showing that synthetic data can be more effective than training from scratch alone. Nevertheless, real data remains essential for achieving robust performance and combining artificial and real data is likely to achieve the best results.

\begin{figure}[h!]
    \centering
    \includegraphics[trim={0 10 0 20}, clip, width=0.76\linewidth]{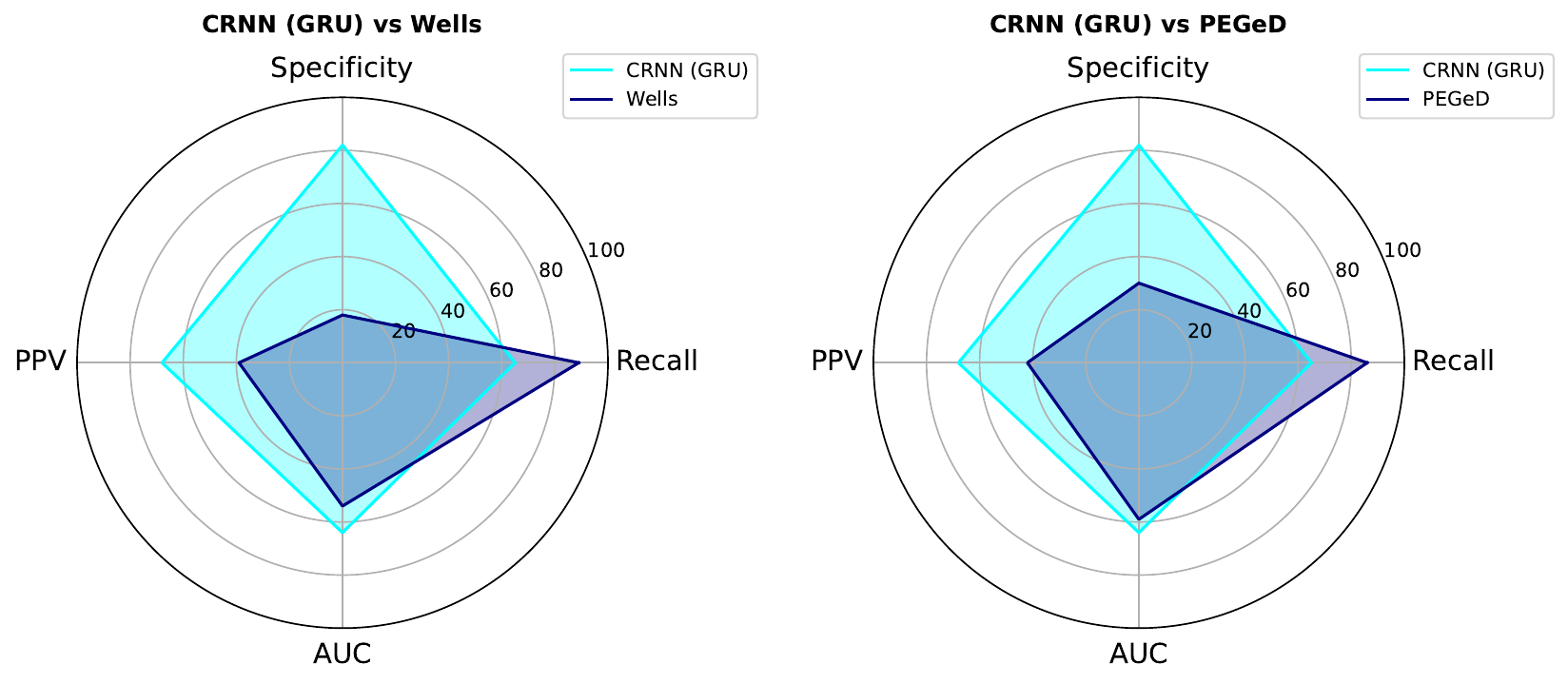}
    \caption{CRNN\textsubscript{GRU} diagnostic performance against Wells (\textbf{left}) and PEGeD (\textbf{right}).}.
    \label{fig:clinicalvsai}
\end{figure}

Finally, we evaluated our best model, CRNN\textsubscript{GRU} (pretrained on PTB-XL) on the PE-HSM test set against two clinical scores for PE diagnosis, Wells score and PEGeD \cite{VALENTESILVA2023643}, as shown in Figure \ref{fig:clinicalvsai}. Although both clinical scores exhibit high sensitivity, the CRNN\textsubscript{GRU} achieves superior area under the curve (AUC), specificity, and positive predictive value (PPV), highlighting the potential of deep learning to enhance PE detection even in resource-limited settings.

\begin{comment}
\begin{figure}[htbp]
    \centering
    % First image
    \begin{minipage}{0.45\textwidth} % Adjust width
        \centering
        \includegraphics[width=0.9\linewidth, trim=50 0 50 0, clip]{images/CRNN-PEG.pdf}

        %\caption{CRNN\textsubscript{GRU} and PEGeD.}
        \label{fig:crnn-peg}
    \end{minipage}
    \hfill
    % Second image
    \begin{minipage}{0.45\textwidth} % Adjust width
        \centering
        \includegraphics[width=0.9\linewidth, trim=50 0 50 0, clip]{images/CRNN-Wells.pdf}
    
        %\caption{CRNN\textsubscript{GRU} and Wells.}
        \label{fig:crnn_wells}
    \end{minipage}
    
    \caption{CRNN\textsubscript{GRU} diagnostic performance against PEGeD (\textbf{Left}) and Wells (\textbf{Right}), which are some of the best clinical scores performed by doctors to detect PE \cite{VALENTESILVA2023643}.}
    \label{fig:clinicalvsai}
\end{figure}
\end{comment}

\section{Conclusions}

In this study, we propose a novel methodology for PE detection using ECG signals, detailing key steps such as input length selection, model-specific normalization, and implementation best practices. Our approach is evaluated across three benchmark datasets, followed by transfer learning to enhance performance on a PE-specific dataset. Results show that ECGs alone can be enough,  providing strong classification baselines and supporting clinical decision-making.

\section{Acknowledgements}
This work was supported by project nº 62 - Center for Responsible AI ref. C628696807-00454142, financed by the Recovery and Resilience Facility - Component 5, included in the NextGenerationEU funding program, by project PRELUNA, grant
PTDC/CCIINF/4703/2021, and by national funds through FCT, Fundação para a Ciência e a Tecnologia, under the project UIDB/50021/2020 (DOI:10.54499/UIDB/50021/2020). We also extend our thanks to Beatriz Valente Silva, Miguel Nobre Menezes, and Fausto J. Pinto for their research contributions and for granting us access to the PE-HSM dataset.

\bibliographystyle{splncs04}
\bibliography{Mybibliography}

\end{document}